\pdfoutput=1

\documentclass[11pt]{article}

\usepackage{acl}

\usepackage{makecell}
\usepackage{times}
\usepackage{amsmath}
\usepackage{amssymb}
\usepackage{amsfonts}
\usepackage{graphicx}
\usepackage{latexsym}
\usepackage{CJKutf8}
\usepackage{amsfonts}
\usepackage{amsmath}
\usepackage{mathtools}
\usepackage{multicol}
\usepackage{multirow}
\usepackage{amssymb}

\usepackage[T1]{fontenc}

\usepackage[utf8]{inputenc}

\usepackage{microtype}

\title{Unified BERT for Few-shot Natural Language Understanding}

\author{Junyu Lu$^{1,2,*}$, Ping Yang$^{1}$, Ruyi Gan$^{1}$, Jing Yang$^{1}$, Jiaxing Zhang$^{1}$
\\
         $^{1}$International Digital Economy Academy \\ $^{2}$South China University of Technology\\
         \texttt{lujunyu@idea.edu.cn}\\
         \texttt{yangping@idea.edu.cn} \\}

\begin{document}
\maketitle
\begin{abstract}
Even as pre-trained language models share a semantic encoder, natural language understanding suffers from a diversity of output schemas. In this paper, we propose UBERT, a unified bidirectional language understanding model based on BERT framework, which can universally model the training objects of different NLU tasks through a biaffine network. Specifically, UBERT encodes prior knowledge from various aspects, uniformly constructing learning representations across multiple NLU tasks, which is conducive to enhancing the ability to capture common semantic understanding. By using the biaffine to model scores pair of the start and end position of the original text, various classification and extraction structures can be converted into a universal, span-decoding approach. Experiments show that UBERT  win the first price in the 2022 AIWIN - World Artificial Intelligence Innovation Competition, Chinese insurance few-shot multi-task track, and realizes the unification of extensive information extraction and linguistic reasoning tasks. 
\end{abstract}

{
\renewcommand{\thefootnote}{\fnsymbol{footnote}}
\footnotetext[1]{Part of this work was done when Junyu Lu interned at International Digital Economy Academy.}
}

\section{Introduction}
Language models (LMs) \cite{devlin2019bert, liu2019roberta, radford2019language, joshi2020spanbert, lewis2020bart, yamada2020luke} have been shown to provide rich semantic support and effectively transfer to various natural language understanding (NLU) downstream tasks \cite{wang2018glue}, including text classification, question answering and information extraction. Pre-trained LMs can learn contextualize text representations by formulating self-supervised training objectives on large amounts of unlabeled text data. However, during the subsequent re-training or fine-tuning of the conjoined model, such approaches construct task-specific decoding structures, each of which relies on an inductive bias related to the object of target task. For question answering, a task-specific span-decoder is often used to extract the span of the answers from the input text \cite{xiong2016dynamic}. For text classification, a task-specific classification layer with assigned categories is typically exploited instead. For named entity recognition, a CRF decoder is often used to retrieve the boundary and category of the entities from the input text. These task-specific inductive biases will lead to jumbled architectures, dedicated models and exclusive knowledge systems for different NLU tasks.

Recent studies have attempted to unify various natural language processing (NLP) tasks using generative LMs based on Transformer architecture. For example, T5 \cite{raffel2020exploring} unifies the training objective and model architecture by converting the input and output into the text paradigm and utilizing task-specific prefixes to distinguish different NLP tasks. UIE \cite{lu2022unified} designs a unified text-to-structure generation architecture that can universally model different information extraction (IE) tasks by encoding heterogeneous IE structures into a uniform representation via a structural extraction language. Due to the consistency objectives of downstream tasks and pre-training, these pre-trained LMs perform well on few-shot and zero-shot scenario. However, such generative pre-trained LMs are influenced by the variability of verbalizers \cite{schick2021exploiting}, and different verbalizer design proposals may lead to unstable task results and exacerbate negative transfer phenomena. 

\begin{figure}[t]
\begin{center}
\centerline{\includegraphics[width=0.95 \linewidth]{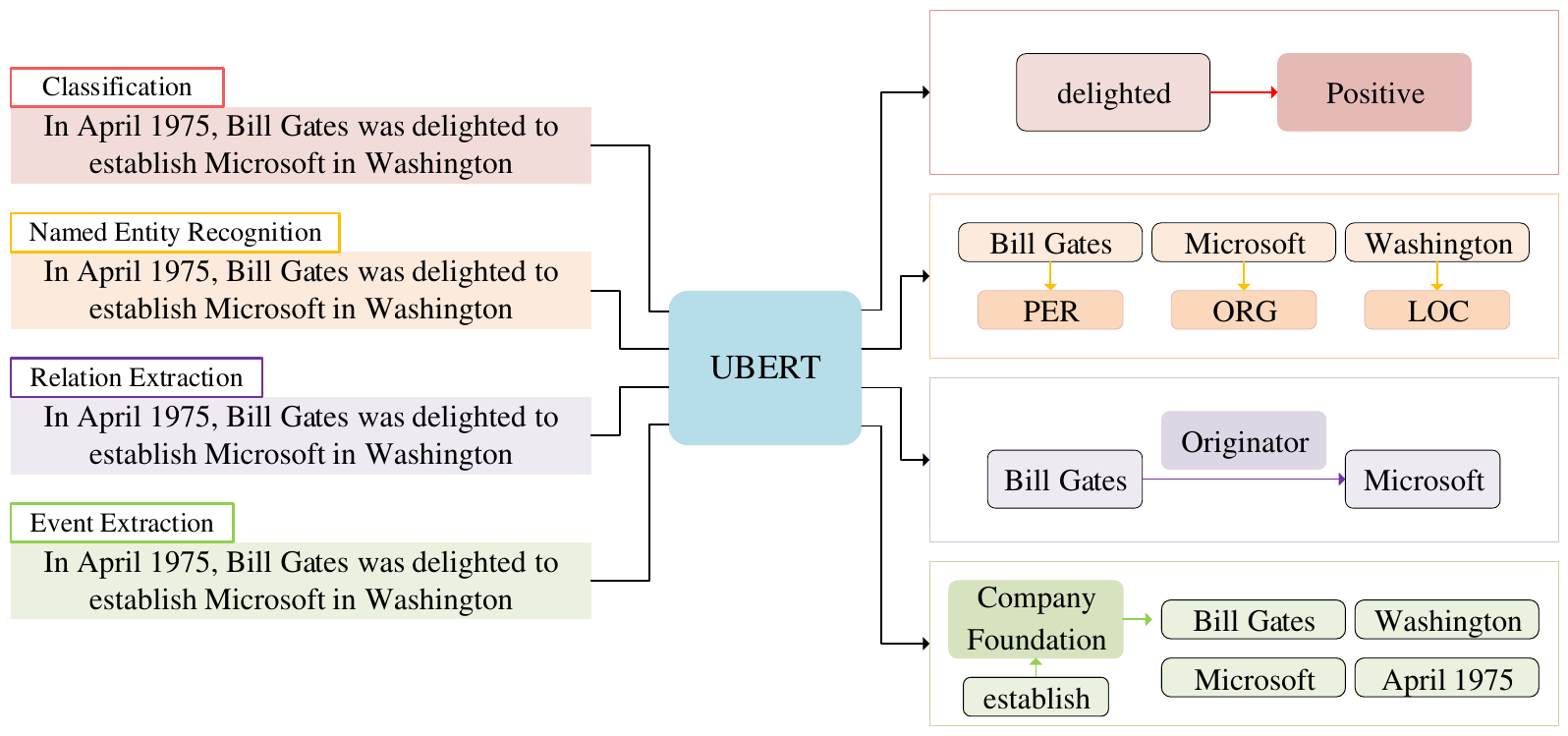}}
\vskip -0.1in
\caption{Task-specialized Natural Language Understanding.}\label{fig:tasks}
\end{center}
\vskip -0.4in
\end{figure}

 \begin{figure*}[t]
\begin{center}
\centerline{\includegraphics[width=0.95 \linewidth]{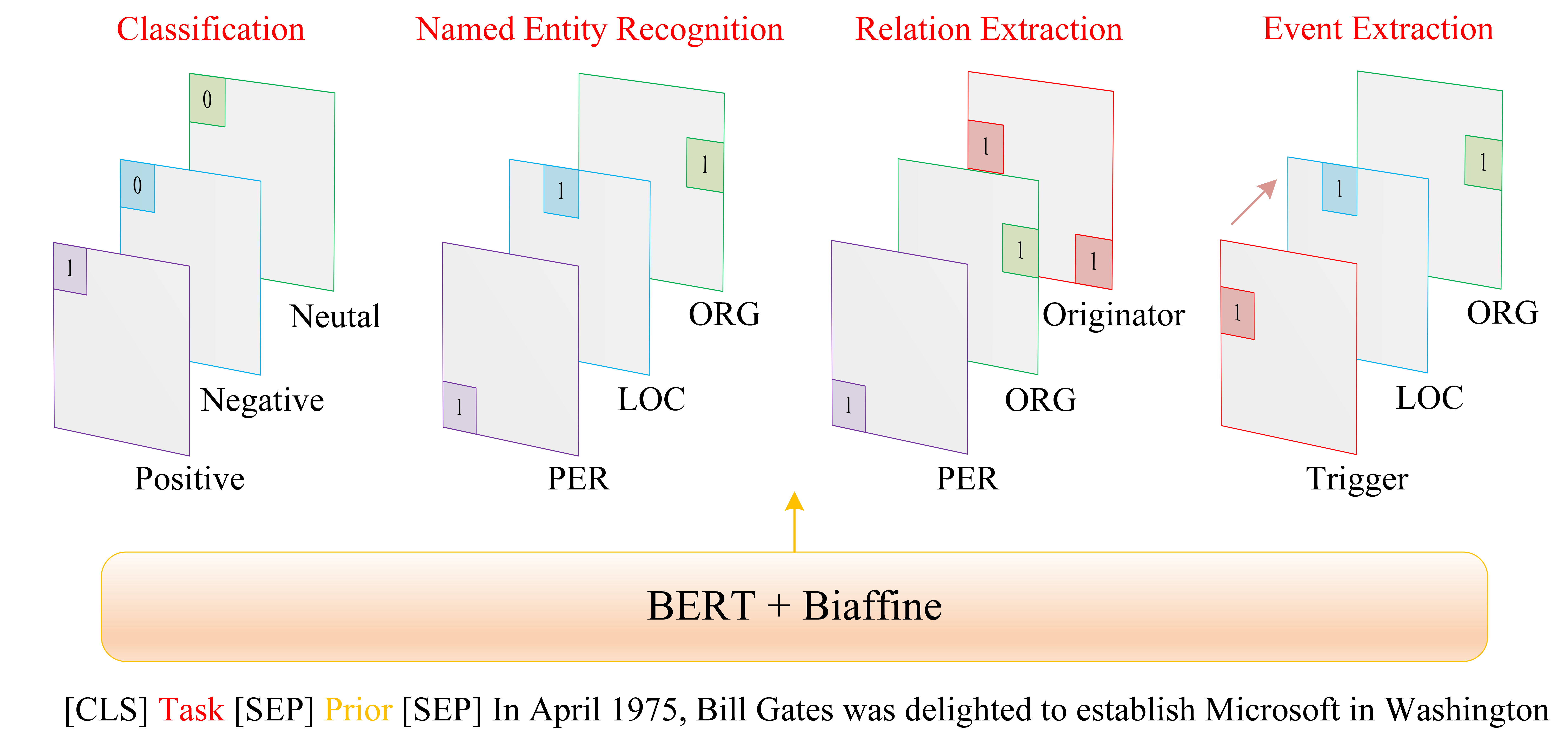}}
\vskip -0.1in
\caption{The overall architecture of UBERT.}\label{fig:ubert_all}
\end{center}
\vskip -0.4in
\end{figure*}

As shown in Figure \ref{fig:tasks}, fundamentally, We consider that all NLU tasks can be modeled as the combination of span-extraction and label classification. Specifically, each span can be an entity, a special token, a mentioned augment or a relation semantic. For example, a span represents the head and tail position of an entity in named entity recognition task, while in text classification or emotion analysis tasks, a span is the special token "[CLS]". Particularly, in relation extraction task, after locating the head and tail positions of the involved entity pairs, the span of their intersections are used to represent the virtual concept of the relationship.

In this case, instead of generating the results sequentially, We are able to encode the text using BERT and decode the hidden output by a span-extraction architecture to obtain arbitrary task object. The overall architecture of our model is shown in Figure \ref{fig:ubert_all}. Firstly, in order to effectively distinguish different NLU tasks and collaboratively learn general representation abilities from various aspects, we construct the task-specific prefixes and auxiliary label information in the text, which can be uniformly fed into the BERT architecture. Then, to generate uniform objects adaptively from hidden sequences, we decompose the sequence decoding into several atomic transformation elements: 1) \textbf{\textit{Structure Table}}, we adopt a biaffine structure \cite{yu2020named} for decoding and obtain a two-dimensional feature map named structure table, where the first dimension represents the head position and the second dimension represents the tail position. For a task-specific sentence, according to the specified prior knowledge, a structure table can be used to extract the demanded information, such as semantics, entities and relations. 2) \textbf{\textit{Locating Designator}}, as we obtain the structure table, we parse the value of all points, and those points that indicate the extraction target are called locating designator. We leverage the correspondence and interleaving relationships between the head and tail of each token to capture locating designators, each of which can be used for label prediction. To learn common NLU abilities for UBERT, we pre-train UBERT on large-scale structured labeled datasets mined from accessible web sources. The large-scaled pre-trained UBERT provides solid semantic support and knowledge sharing capabilities, which enable it to quickly adapt to new NLU scenarios.

We conduct experiments on 14 datasets of 7 main NLU tasks, which including mainly two domains: classification and information extraction.

The main contributions of our paper are:

1) We propose UBERT, a unified text-to-structure extraction architecture that can universally model various NLU tasks as the combination of span-extraction and label classification, adaptively generate the targeted structure through span-extraction and learn general comprehension from task-specific knowledge.

2) We design a span-based biaffine decoding network, which organize the correspondence and interleaving relations between the head and tail of each token via the structure table and locating designator to control which information points are associated.

3) We pre-train a large-scale text-to-structure extraction model via a unified pre-training process on structured labelled datasets of various NLU tasks. To the best of our knowledge, this is the first text-to-structure pre-trained extraction model based on bidirectional language understanding model, which is conducive to promoting unified NLU researches.

\section{Related Work}

In the early stage, masked pre-trained language models were used to provide contextualized representations, such as BERT \cite{devlin2019bert} and RoBERTa \cite{liu2019roberta}, while generative pre-trained language models performed well in text generation, question answering and knowledge reasoning tasks, such as GPT \cite{radford2018improving} and BART \cite{lewis2020bart}. However, when fine-tuning these pre-trained models in downstream tasks, we must construct the task-specific decoding layer over the models.

Recent studies have focused on building and pre-training a unified model to carry out various NLP tasks, which is beneficial for knowledge sharing and saving the training costs for independent models. \cite{dong2019unified} proposed UniLM, a single Transformer LM that uses the shared parameters and architecture for different types of LMs, alleviating the need of separately training and hosting multiple LMs. \cite{aghajanyan2021muppet} proposed an additional large-scale learning stage named pre-finetuning, which introduced a noval multi-task training scheme for effective learning at scale. Furthermore, several studies have been undertaken to design uniform input paradigms and decoding structures to universally model various NLP tasks. \cite{keskar2019unifying} proposed treating question answering, text classification, and regression as span-extractive tasks. \cite{raffel2020exploring} proposed a text-to-text transfer Transformer, converting all NLP tasks as the generation task by adding a task-specific prefix in the raw text. \cite{lu2022unified} designed a text-to-structure framework to uniformly model the information extraction task, which encodes different extraction structures via a structured extraction language, adaptively generates target extractions via a schema-based prompt mechanism. Different from the above approaches, we aim to utilize an extractive structure to model NLU tasks uniformly.

\section{Unified Span Extraction for Universal Natural Language Understanding}
Natural language understanding can be broadly divided into two directions: classification and information extraction, both of which can be formulated as span-extraction problems. This paper aims to uniformly model the span-extraction via a biaffine network, which signifies that different end-to-end transformations will share the same underlying semantic encoding and perform different structure decoding operations. Generally, there are two main challenge to unify the architecture of different NLU tasks. Firstly, for a unified extractive structure, the model needs to be aware of what to extract and which to identify. Secondly, due to the diversity of NLU tasks, different target structures request the structure tables to be characterized to corresponding meanings, such as emotion, entity and relationship.

To address the first issue, we design a uniform schema to construct the input sentence. Formally, given a specific task name $t$, we denote the input texts as $S=\{s_1,s_2,\ldots,s_n\}$ and the pre-defined categories set as $C=\{c_1,c_2,\ldots,c_m\}$. For emotion analysis, $c_j$ represents a sentimental word such as positive and negative. For named entity recognition, $c_j$ represents an entity category such as person, location and organization. For relation extraction, $c_j$ represents a triple contained the head entity, relation type and tail entity, such as \{PER, Originator, ORG\}. For event extraction, $c_j$ represents a combination of event type and argument role, such as \{enterprise establishing, time\} and \{attack, victim\}. The category information provides rich prior knowledge for the training target, effectively guiding the prediction of locating designators. We match each sentence with $m$ tags individually, and each input can be formulated as a uniform schema by stacking $m$ pieces of $x_{ij}=\{\mathrm{[task]}\ t\ \mathrm{[category]}\ c_j\ \mathrm{[text]}\ s_i\}$, where $X_{i}=\{x_{i1},x_{i2},\ldots,x_{im}\}$.

\subsection{Task-specific Span-extraction Decoding}

Firstly, after obtaining the word representations $x$ from BERT, we apply two separate feed-forward neural network (FFN) to create different representations ($h_s / h_e$) for the start/end of the spans in structure table. Using different representations for the start/end of the spans allow the model to learn the information independence the start/end of the spans. Finally, we employ the biaffine network  over the sentence to create a two-dimensional structure table $Score(s,e) \in \mathbb{R}^{l \times l \times 1}$ for scoring each predicted point, while $l$ is the length of the sentence. We calculate the structure table using \textbf{einsum} to sum out the $i$-th row and $j$-th column:
\begin{equation}
\begin{array}{c}
h_{s}=\operatorname{FFN_{s}}(x)\\
h_{e}=\operatorname{FFN_{e}}(x)\\
\operatorname{Score}(s, e)=\operatorname{\mathbf{einsum}}\left(\text{``} l d, d z d, d l\text{''}, h_{s}, \mathrm{U}, h_{e}\right)
\end{array}
\end{equation}

Then, we propose treating classification, named entity recognition, relation extraction, and event extraction as span-extractive tasks. We design a common decoding structure for four tasks, with different representational meaning and corresponding entries of the locating designators. Note that each target point in the structure table is a binary value, we use locating designator to indicate that position is activated with specific semantic, and use zero to indicate that the position is overruled or meaningless. We will describe how to determine the position and meaning of locating designators in specific tasks in the following.

\paragraph{Classification}
For classification task, we treat the intersection of the head position to the tail position of the special token "[CLS]" in the two-dimensional structure table as an locating designator. The predicted point indicates whether the text belongs to the category $c_j$ in the input scheme, the value of the locating designator will be set to one if so, otherwise it will be set to zero. Meanwhile, all other positions in the table will also be set to zero.

\paragraph{Named Entity Recognition}
Similar to classification task, for each entity, we use its start position and end position in the raw text as the coordinates of its first dimension and second dimension in the structure table, respectively. In a structure table, we gather all entities that match the category in the input schema and set the corresponding position to one.

\paragraph{Relation Extraction}
The decoding of relation extraction task can be regarded as the coupling of entity recognition and relation classification. In this case, different from classification and named entity recognition tasks that use a structure table to decode a category, for each relation type, we need two extra structure tables to decode the entities. Specifically, the first two structure tables obtain the head entity and tail entity as the decoding of named entity recognition, whose positions can be represented as $(s_h,e_h)$ and $(s_t,e_t)$. Then, in the third structure table, we aims at determining two locating designators based on the interlaced information of the start and position of the head and tail entities, with the corresponding positions $(s_h,s_t)$ and $(e_h,e_t)$ are set to one.

\paragraph{Event Extraction}
Different with the above tasks, we treat the event extraction be as a two-stage named entity recognition task, where the trigger is determined in the first stage and the event augments are extracted in the second stage. Formally, to decode a event type entirely, we first use a piece of event input with the event type as the prior knowledge, and a structure table is first used to define the trigger. Then, we combine the trigger with each of the corresponding augment type as the input schema and append extra structure tables for decoding each event augment as named entity recognition.

\paragraph{Training Objective}
In the training phase, the $m$-piece texts of each normalized input are stacked and packed into a complete unit as the input to BERT and we uniformly use multi-label classification as the training target for different NLU tasks. Instead of using sigmoid to compute the loss for all positions of each table individually, we introduce binary cross-entropy (BCE) as the loss function. which can correlate the dependencies of different information points. For an input text $X_{i}$ and output $\hat{y}_{i}$, we flatten the three-dimension structure tables into a one-dimension vector, and the corresponding output $y_{i}$ is a Boolean vector with the locations of information points set to 1 and others set to 0. The loss can be calculated as following:
\begin{equation}
L=-\sum_{i} y_{i} \log \left(\sigma\left(\hat{y}_{i}\right)\right)+\left(1-y_{i}\right) \log \left(1-\sigma\left(\hat{y}_{i}\right)\right)
\end{equation}

\section{Experiments}
We conduct extensive experiments on NLU datasets, and win the first price in the 2022 AIWIN - World Artificial Intelligence Innovation Competition, Chinese insurance few-shot multi-task track, which aims to complete all NLU tasks with a single model.

\bibliography{anthology,custom}

\begin{thebibliography}{16}
\expandafter\ifx\csname natexlab\endcsname\relax\def\natexlab#1{#1}\fi

\bibitem[{Aghajanyan et~al.(2021)Aghajanyan, Gupta, Shrivastava, Chen,
  Zettlemoyer, and Gupta}]{aghajanyan2021muppet}
Armen Aghajanyan, Anchit Gupta, Akshat Shrivastava, Xilun Chen, Luke
  Zettlemoyer, and Sonal Gupta. 2021.
\newblock Muppet: Massive multi-task representations with pre-finetuning.
\newblock In \emph{Proceedings of the 2021 Conference on Empirical Methods in
  Natural Language Processing}, pages 5799--5811.

\bibitem[{Devlin et~al.(2019)Devlin, Chang, Lee, and
  Toutanova}]{devlin2019bert}
Jacob Devlin, Ming-Wei Chang, Kenton Lee, and Kristina Toutanova. 2019.
\newblock Bert: Pre-training of deep bidirectional transformers for language
  understanding.
\newblock In \emph{Proceedings of the 2019 Conference of the North American
  Chapter of the Association for Computational Linguistics: Human Language
  Technologies, Volume 1 (Long and Short Papers)}, pages 4171--4186.

\bibitem[{Dong et~al.(2019)Dong, Yang, Wang, Wei, Liu, Wang, Gao, Zhou, and
  Hon}]{dong2019unified}
Li~Dong, Nan Yang, Wenhui Wang, Furu Wei, Xiaodong Liu, Yu~Wang, Jianfeng Gao,
  Ming Zhou, and Hsiao-Wuen Hon. 2019.
\newblock Unified language model pre-training for natural language
  understanding and generation.
\newblock \emph{Advances in Neural Information Processing Systems}, 32.

\bibitem[{Joshi et~al.(2020)Joshi, Chen, Liu, Weld, Zettlemoyer, and
  Levy}]{joshi2020spanbert}
Mandar Joshi, Danqi Chen, Yinhan Liu, Daniel~S Weld, Luke Zettlemoyer, and Omer
  Levy. 2020.
\newblock Spanbert: Improving pre-training by representing and predicting
  spans.
\newblock \emph{Transactions of the Association for Computational Linguistics},
  8:64--77.

\bibitem[{Keskar et~al.(2019)Keskar, McCann, Xiong, and
  Socher}]{keskar2019unifying}
Nitish~Shirish Keskar, Bryan McCann, Caiming Xiong, and Richard Socher. 2019.
\newblock Unifying question answering, text classification, and regression via
  span extraction.
\newblock \emph{arXiv preprint arXiv:1904.09286}.

\bibitem[{Lewis et~al.(2020)Lewis, Liu, Goyal, Ghazvininejad, Mohamed, Levy,
  Stoyanov, and Zettlemoyer}]{lewis2020bart}
Mike Lewis, Yinhan Liu, Naman Goyal, Marjan Ghazvininejad, Abdelrahman Mohamed,
  Omer Levy, Veselin Stoyanov, and Luke Zettlemoyer. 2020.
\newblock Bart: Denoising sequence-to-sequence pre-training for natural
  language generation, translation, and comprehension.
\newblock In \emph{Proceedings of the 58th Annual Meeting of the Association
  for Computational Linguistics}, pages 7871--7880.

\bibitem[{Liu et~al.(2019)Liu, Ott, Goyal, Du, Joshi, Chen, Levy, Lewis,
  Zettlemoyer, and Stoyanov}]{liu2019roberta}
Yinhan Liu, Myle Ott, Naman Goyal, Jingfei Du, Mandar Joshi, Danqi Chen, Omer
  Levy, Mike Lewis, Luke Zettlemoyer, and Veselin Stoyanov. 2019.
\newblock Roberta: A robustly optimized bert pretraining approach.
\newblock \emph{arXiv preprint arXiv:1907.11692}.

\bibitem[{Lu et~al.(2022)Lu, Liu, Dai, Xiao, Lin, Han, Sun, and
  Wu}]{lu2022unified}
Yaojie Lu, Qing Liu, Dai Dai, Xinyan Xiao, Hongyu Lin, Xianpei Han, Le~Sun, and
  Hua Wu. 2022.
\newblock Unified structure generation for universal information extraction.
\newblock In \emph{Proceedings of the 60th Annual Meeting of the Association
  for Computational Linguistics (Volume 1: Long Papers)}, pages 5755--5772.

\bibitem[{Radford et~al.()Radford, Narasimhan, Salimans, Sutskever
  et~al.}]{radford2018improving}
Alec Radford, Karthik Narasimhan, Tim Salimans, Ilya Sutskever, et~al.
\newblock Improving language understanding by generative pre-training.

\bibitem[{Radford et~al.(2019)Radford, Wu, Child, Luan, Amodei, Sutskever
  et~al.}]{radford2019language}
Alec Radford, Jeffrey Wu, Rewon Child, David Luan, Dario Amodei, Ilya
  Sutskever, et~al. 2019.
\newblock Language models are unsupervised multitask learners.

\bibitem[{Raffel et~al.(2020)Raffel, Shazeer, Roberts, Lee, Narang, Matena,
  Zhou, Li, and Liu}]{raffel2020exploring}
Colin Raffel, Noam Shazeer, Adam Roberts, Katherine Lee, Sharan Narang, Michael
  Matena, Yanqi Zhou, Wei Li, and Peter~J Liu. 2020.
\newblock Exploring the limits of transfer learning with a unified text-to-text
  transformer.
\newblock \emph{Journal of Machine Learning Research}, 21:1--67.

\bibitem[{Schick and Sch{\"u}tze(2021)}]{schick2021exploiting}
Timo Schick and Hinrich Sch{\"u}tze. 2021.
\newblock Exploiting cloze-questions for few-shot text classification and
  natural language inference.
\newblock In \emph{Proceedings of the 16th Conference of the European Chapter
  of the Association for Computational Linguistics: Main Volume}, pages
  255--269.

\bibitem[{Wang et~al.(2018)Wang, Singh, Michael, Hill, Levy, and
  Bowman}]{wang2018glue}
Alex Wang, Amanpreet Singh, Julian Michael, Felix Hill, Omer Levy, and Samuel
  Bowman. 2018.
\newblock Glue: A multi-task benchmark and analysis platform for natural
  language understanding.
\newblock In \emph{Proceedings of the 2018 EMNLP Workshop BlackboxNLP:
  Analyzing and Interpreting Neural Networks for NLP}, pages 353--355.

\bibitem[{Xiong et~al.(2016)Xiong, Zhong, and Socher}]{xiong2016dynamic}
Caiming Xiong, Victor Zhong, and Richard Socher. 2016.
\newblock Dynamic coattention networks for question answering.
\newblock \emph{arXiv preprint arXiv:1611.01604}.

\bibitem[{Yamada et~al.(2020)Yamada, Asai, Shindo, Takeda, and
  Matsumoto}]{yamada2020luke}
Ikuya Yamada, Akari Asai, Hiroyuki Shindo, Hideaki Takeda, and Yuji Matsumoto.
  2020.
\newblock Luke: Deep contextualized entity representations with entity-aware
  self-attention.
\newblock In \emph{Proceedings of the 2020 Conference on Empirical Methods in
  Natural Language Processing (EMNLP)}, pages 6442--6454.

\bibitem[{Yu et~al.(2020)Yu, Bohnet, and Poesio}]{yu2020named}
Juntao Yu, Bernd Bohnet, and Massimo Poesio. 2020.
\newblock Named entity recognition as dependency parsing.
\newblock In \emph{Proceedings of the 58th Annual Meeting of the Association
  for Computational Linguistics}, pages 6470--6476.

\end{thebibliography}
\bibliographystyle{acl_natbib}

\appendix

\section{Example Appendix}
\label{sec:appendix}

This is an appendix.

\end{document}